\documentclass[conference]{IEEEtran}
\IEEEoverridecommandlockouts
\usepackage[letterpaper,top=0.75in,bottom=0.75in,left=0.75in,right=0.75in,heightrounded]{geometry}
\usepackage{cite}
\usepackage{amsmath,amssymb,amsfonts}
\usepackage{graphicx}
\usepackage{textcomp}
\usepackage{xcolor}
\usepackage{hyperref}
\usepackage{algorithm}
\usepackage{algorithmicx}
\usepackage{algpseudocode}
\usepackage{stfloats}

\algrenewcommand\algorithmicrequire{\textbf{Input:}}
\algrenewcommand\algorithmicensure{\textbf{Output:}}

\def\BibTeX{{\rm B\kern-.05em{\sc i\kern-.025em b}\kern-.08em
    T\kern-.1667em\lower.7ex\hbox{E}\kern-.125emX}}
\begin{document}

\title{\vspace*{0.25in}Towards Autonomous Instrument Tray Assembly for Sterile Processing Applications}

\author{
Raghavasimhan Sankaranarayanan$^{1}$,
Paul Stuart$^{2}$,
Nicholas Ahn$^{3}$,
Arno Sungarian$^{4}$,
and Yash Chitalia$^{1}$
\thanks{
$^{1}$Department of Mechanical Engineering, University of Louisville, Louisville, KY, USA. 
$^{2}$Sterile Processing Department, Saint Vincent's Hospital, Worcester, MA, USA.
$^{3}$Department of Orthopedic Surgery, University of Louisville, Louisville, KY, USA.
$^{4}$Department of Neurological Surgery, University of Massachusetts, Worcester, MA, USA. 
\texttt{r0sank01@louisville.edu}}
}

\maketitle
\begin{abstract}
The Sterile Processing and Distribution (SPD) department is responsible for cleaning, disinfecting, inspecting, and assembling surgical instruments between surgeries. Manual inspection and preparation of instrument trays is a time-consuming, error-prone task, often prone to contamination and instrument breakage. 
In this work, we present a fully automated robotic system that sorts and structurally packs surgical instruments into sterile trays, focusing on automation of the SPD assembly stage.
A custom dataset comprising 31 surgical instruments and 6,975 annotated images was collected to train a hybrid perception pipeline using YOLO12 for detection and a cascaded ResNet-based model for fine-grained classification.
The system integrates a calibrated vision module, a 6-DOF Stäubli TX2-60L robotic arm with a custom dual electromagnetic gripper, and a rule-based packing algorithm that reduces instrument collisions during transport. 
The packing framework uses 3D printed dividers and holders to physically isolate instruments, reducing collision and friction during transport. 
Experimental evaluations show high perception accuracy and statistically significant reduction in tool-to-tool collisions compared to human-assembled trays. 
This work serves as the scalable first step toward automating SPD workflows, improving safety, and consistency of surgical preparation while reducing SPD processing times.
\end{abstract}


\section{Introduction}
\begin{figure}[htb]
    \centering
    \includegraphics[width=0.9\columnwidth]{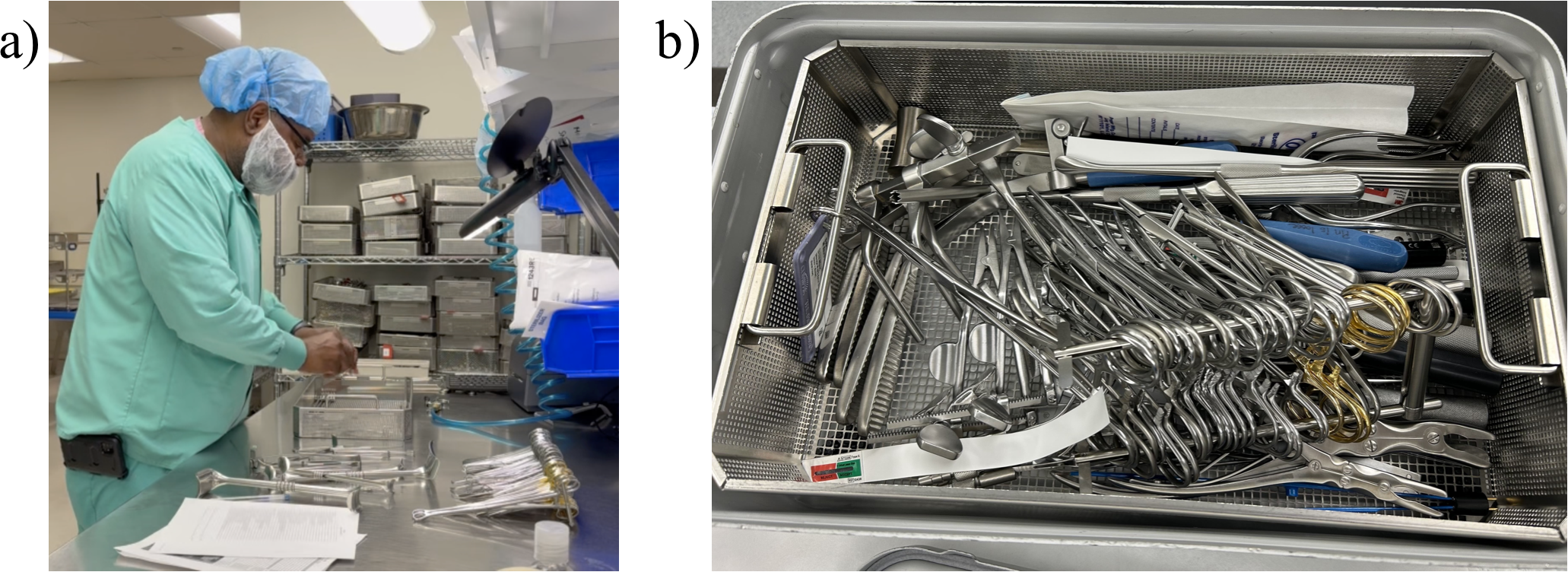}
    \caption{a) An SPD technician inspecting and assembling a tray. b) A typical freshly opened tray in the Operating Room}
    \label{fig:clutter}
\end{figure}

Sterile Processing and Distribution (SPD) Technicians are responsible for cleaning and inspecting surgical instruments, assembling instruments within sterile trays, and sterilizing trays prior to clinical use~\cite{HOLMES2020443} (typical inspection and assembly by an SPD technician is shown in \autoref{fig:clutter}a). 
These are critical to pre-operative workflows, yet remain highly manual and time-consuming~\cite{hung2020using}. 
In high-stakes environments such as hospitals, technicians often face significant workload and time pressure, which can lead to inconsistencies in tray assembly, misplaced instruments, and increased risk of damage or contamination~\cite{Manikonda2025}. 
Inefficient or cluttered tray layouts (shown in \autoref{fig:clutter}b) can delay surgical procedures, elevate cognitive load on operating room (OR) staff, and reduce the longevity of expensive surgical instruments due to tool-to-tool collisions during transportation or sterilization.

\begin{figure}[htb]
    \centering
    \includegraphics[width=0.85\columnwidth]{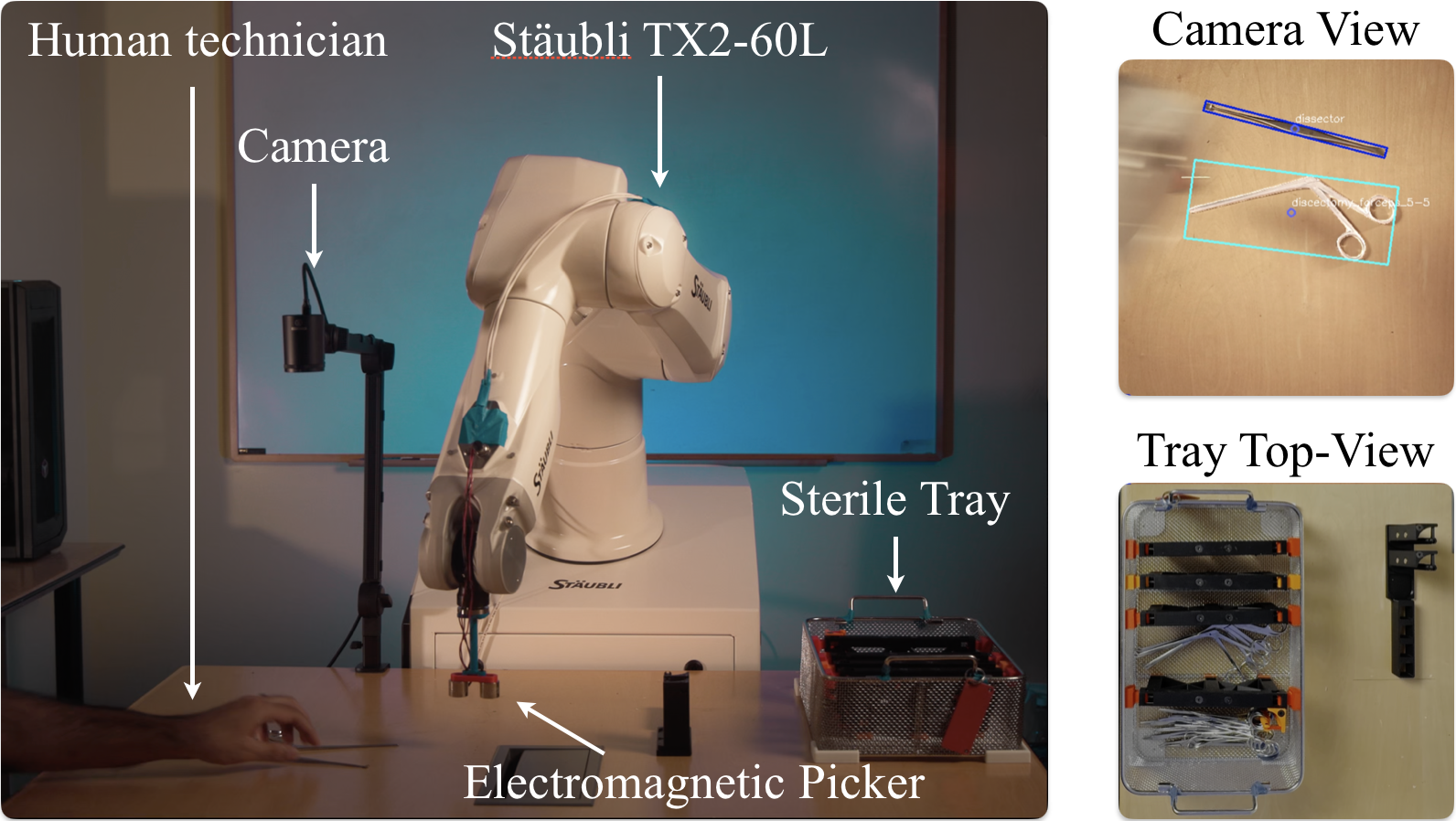}
    \caption{The robotic system working with a human, showing what the robot sees and the top view of the tray during the assembly process}
    \label{fig:hri_setup}
\end{figure}

Automating this process can substantially improve efficiency and consistency while reducing the burden on SPD personnel. 
In this work, we address one critical step of the SPD workflow - the assembly and organization of instruments into the sterile tray. 
We propose a robotic system that performs end-to-end instrument sorting and structured packing using vision-based perception, motion planning, and custom-designed hardware. 
The system integrates a 6-DOF Stäubli TX2-60L robotic manipulator~\cite{staubli} with a dual electromagnetic gripper and a calibrated top-down vision system for real-time detection, classification, and structured placement of surgical instruments in the sterile tray. 
As part of the assembly, the robot inserts custom 3D printed dividers and holders to physically isolate instruments, ensuring reproducible spatial layouts that reduce collisions and simplify post-sterilization retrieval by OR nurses. 
\autoref{fig:hri_setup} shows our proposed collaborative setup where a human technician inspects instruments while the robot autonomously arranges them in the tray.

Given the subtle variations and complex geometries of surgical instruments, standard object detection pipelines struggle to differentiate instruments, especially with smaller data. 
To address this, we developed a custom dataset and a hybrid perception framework that combines YOLO12~\cite{tian2025yolov12} for real-time bounding box detection, the Segment Anything Model (SAM)~\cite{kirillov2023segment} for mask extraction and pose estimation, and a custom ResNet-based~\cite{he2016deep} architecture for classification. 
The resulting coordinates are transformed into the world frame using calibrated camera extrinsics. 
An SPD-inspired packing algorithm determines instrument placements that reduce movement and collisions while maintaining a spatial organization consistent with OR conventions.

The proposed system targets hospital sterilization units and OR support workflows where instrument preparation is typically performed under tight time constraints by non-surgical staff. 
By offloading sorting and assembly tasks to a robotic system, hospitals can reduce labor burden, improve reliability, and ensure safer, standardized tray configurations, paving the way for automated verification and inventory management in the future. Our system demonstrates a hybrid human-robot workflow. The technician performs the inspection while the robot autonomously handles sorting and structured packing.

\subsection{Related works}
\label{sec:related_works}
Despite the high importance of this workflow, most existing automation research in surgical robotics focuses on intra-operative assistance rather than pre-operative preparation. As an early step toward automating sterile processing, recent studies have explored robotic systems for instrument recognition, grasping, and manipulation in cluttered environments.

Xu et al.~\cite{xu_vision-guided_2014} used a robotic manipulator with classical edge-based vision to isolate instruments, but their pipeline requires controlled lighting. 
Wu et al.~\cite{wu_coordinated_2019} developed a dual-arm system for grasping using binocular vision, yet their system lacked tray assembly logic. 
Zhou et al.~\cite{zhou_needle_2017} proposed interactive surgical instrument recognition through perception and manipulation in the OR environment, while Kim et al.~\cite{kim_surggrip_2022} introduced a compliant 3D printed gripper optimized for thin instruments; however, it is not an end-to-end system and did not include aspects of tray assembly. 
Li et al.~\cite{li_surgical_2017} employed stereo vision with impedance control for instrument localization, but were limited in tool types. Other works~\cite{deol_artificial_2024, ng_multimodal_2024, lavado_sorting_2018} have advanced perception techniques to handle occlusions and complex instrument geometries. 
There have also been contributions on datasets such as the HOSPI dataset~\cite{long_evaluation_2022}, but none of them provide calibration parameters. 

None of these efforts addresses the automated tray assembly, which involves sorting and securely packing instruments for sterilization and subsequent surgical use. 
Furthermore, visually similar instruments such as straight vs. curved forceps or mosquito vs. crile forceps, remain a challenge for most recognition systems with smaller training data.

We bridge this gap by developing an end-to-end robotic pipeline to detect and assemble surgical instruments into sterile trays with structured layouts. 
The system benefits SPD technicians and OR nurses by reducing processing times, and instrument collisions while enhancing overall workflow efficiency.

In summary, our contributions are as follows:

\begin{itemize}
    \item We introduce a complete pipeline for the automated sorted assembly of surgical instruments using a vision-guided robotic arm and dual electromagnetic gripper.
    \item A custom dataset and a part-aware perception architecture to achieve robust instrument detection and pose estimation.
    \item A generalizable packing-optimization framework that formalizes surgical instrument placement as a ``constrained spatial allocation" problem with collision-reducing objectives while simplifying OR preparation.
\end{itemize}

The remainder of the paper is structured as follows: In section~\ref{sec:method} we detail our methodology including system design (section~\ref{sec:system_design}), vision (section~\ref{sec:vision}), robot manipulation (section~\ref{sec:manipulation}) as well as the packing algorithm (section~\ref{sec:packing}) followed by evaluation in section~\ref{sec:eval}. 
Finally, we conclude and discuss our future directions in section~\ref{sec:future_work}.

\section{Methodology}
\label{sec:method}
\subsection{Proposed System Workflow}
\label{sec:system_design}
The proposed system is designed to collaborate with human technicians, ensuring interpretability and compliance with existing SPD protocols.
In a conventional SPD workflow, after the cleaning process is complete, the technician would first inspect each instrument for any debris, including dust, human tissue residual, blood, etc., as well as for functionality. 
The technician would refer to a list of instruments required for a given procedure and assemble the sterile tray accordingly. 
It is a meticulous process that involves multitasking and is thus prone to errors. 
The proposed system consists of a calibrated camera viewing the stage (where the instruments would be placed by the technician after inspection) and a 6-DOF Robotic arm. 
The camera and the robot are hand-eye calibrated (mean reprojection error $< 0.3$ pixels). 
The system accesses a digital checklist of instruments for each procedure.
It computes the tray assembly beforehand using the algorithm discussed in section~\ref{sec:packing}. 
When the technician places an instrument on the stage after inspection, the robot will place the instrument in the sterile tray in a systematic manner.
If the placed instrument is not in the right order, the robot waits for the technician to place the next instrument before placing the current one, and if the instrument is not part of the requirements, the robot discards it and continues to wait for the next one. 
The robot will take care of placing the instrument in the correct order. 
The technician just inspects the instruments and does not have to worry about requirements or placing them neatly in the tray.

\subsection{Vision}
\label{sec:vision}
\subsubsection{Dataset}
\label{sec:dataset}
\begin{figure}[htb]
    \centering
    \includegraphics[width=\columnwidth]{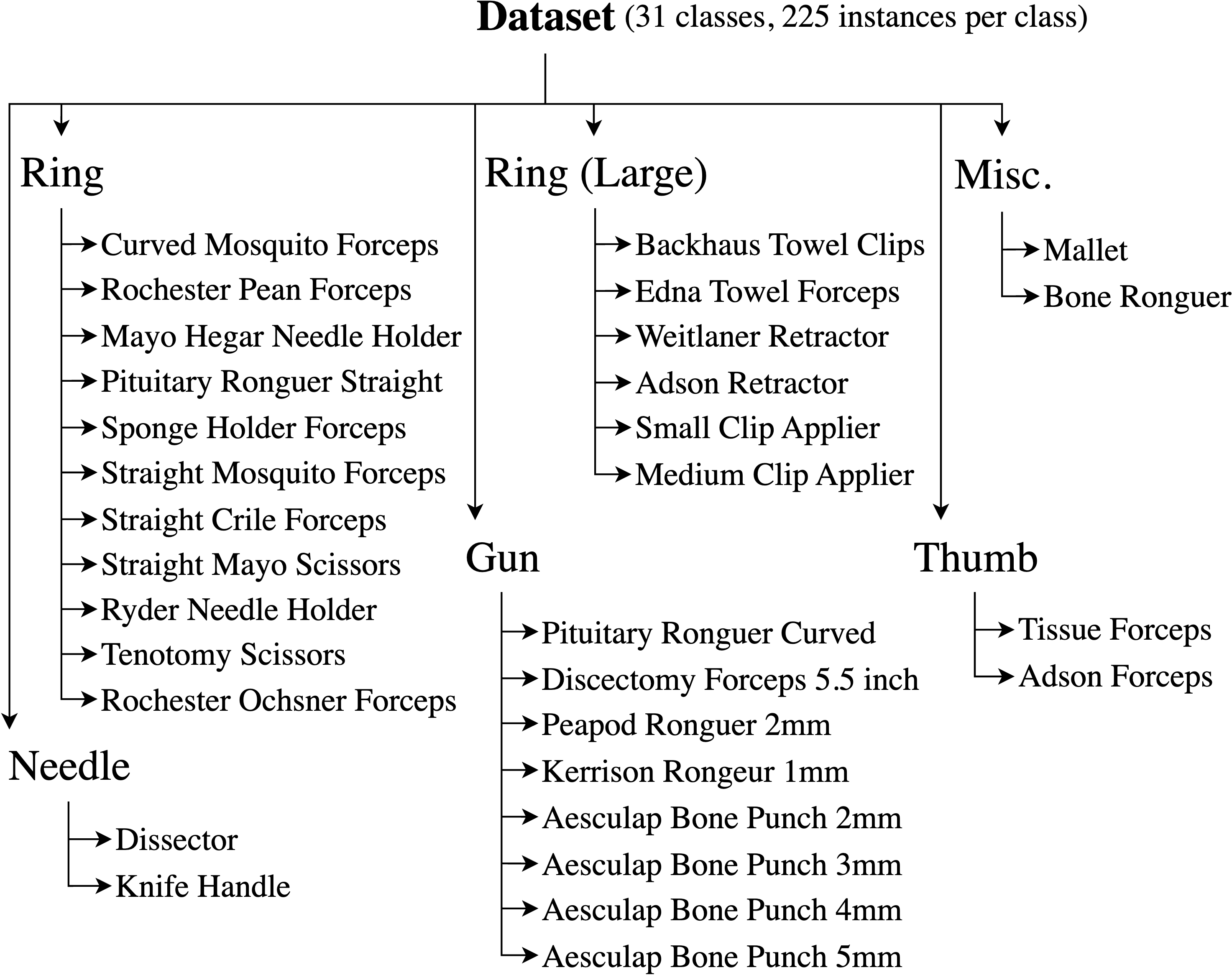}
    \caption{List of instruments in the dataset by the instrument group}
    \label{fig:tools}
\end{figure}
As discussed in section~\ref{sec:related_works}, there are numerous datasets, such as the HOSPI dataset~\cite{long_evaluation_2022}, that are publicly available to train machine learning models. 
However, since we needed the images to be shot with a known camera intrinsic and pose (detailed in section~\ref{sec:detection}), and since we do not have the same set of surgical instruments with us for evaluation, we collected our own data that fits our system. 
We used 31 different surgical instruments in the dataset. 
We collected 250 top-view images per instrument in different lighting conditions, backgrounds, and Z-distances. 
After cleaning the data, we ended up with 225 instances per class, totaling 6975 images, out of which 80\% was used for training and the remaining data was equally split for validation and test. 
One of our goals was to model the subtle feature differences in instruments - not studied in any of the previous works. 
Thus, we chose instruments that include similar features but different sizes, such as the 2mm bone punch vs. the 3mm bone punch, and small differences in features while having the same size, such as the straight vs. curved mosquito forceps. \autoref{fig:tools} shows the list of instruments by their respective instrument group they belong. 

We annotated the instrument bounding boxes in YOLO format. 
The YOLO 12 nano (yolo12n) model~\cite{tian2025yolov12} performs well on small data on the detection task as explained in section~\ref{sec:detection}. 
We leveraged this to speed up annotations. 
We first manually annotated a subset containing 50 instances per class using Roboflow~\cite{roboflow} and trained the yolo12n model. 
We then used this trained model to automate the further annotation process, speeding up to $\sim$50 images per minute.

\subsubsection{Object Detection}
\label{sec:detection}
\begin{figure*}[b!]
    \centering
    \includegraphics[width=0.9\textwidth]{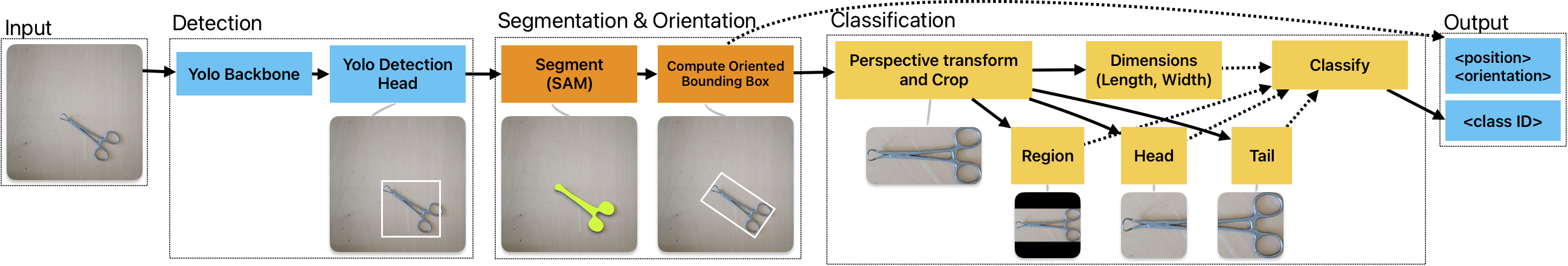}
    \caption{Training and inference pipeline}
    \label{fig:pipeline}
\end{figure*}
Following the related works that largely used flavors of YOLO object detection models~\cite{hussain2023yolo}, we initially collected a small pilot dataset comprising 10 instruments with $\sim$100 instances per class and trained a yolo12n model with an input image size of 480 pixels. 
While the model achieved reliable localization, its classification performance was inconsistent, particularly for instruments with similar geometric profiles and reflective surface characteristics.
To address this limitation, we trained the same dataset using a ResNet-18 classifier.
Compared to YOLO, the ResNet-18 model demonstrated improved separability and reduced ambiguity among visually dissimilar instruments, indicating stronger feature separation at the classification stage.
However, the model continued to struggle with fine-grained distinctions between closely related instruments, resulting in elevated false positives and false negatives.

\begin{figure}[htb]
    \centering
    \includegraphics[width=0.6\columnwidth]{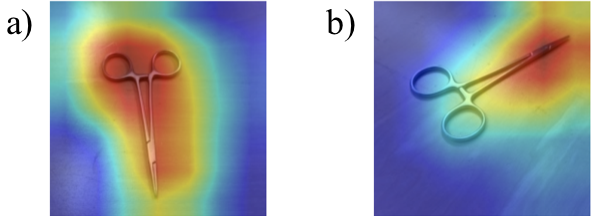}
    \caption{Heat map at layer 4 of ResNet-18. a) failure b) Success}
    \label{fig:gradcam}
\end{figure}

\autoref{fig:gradcam} shows the GradCAM~\cite{selvaraju2016grad} heatmap visualization of the ResNet-18 model viewed at layer 4, showing an example failure case (a) and a success case (b). 
We learned that when the model looks at the tip, the likelihood of predicting the correct class is higher. 
These observations motivated the proposed cascaded Part-Aware ResNet-18 based architecture shown in \autoref{fig:pipeline} (classification), which explicitly targets localized discriminative features for improved fine-grained classification. 
The proposed method decomposes a surgical instrument into its functional parts - head, tail, and full body regions, extracts features from each using ResNet-18 backbones, and concatenates the features along with meta-feature conditioning. 
As mentioned in section~\ref{sec:dataset}, we also have instruments that look similar but only differ in size. 
Both YOLO as well as ResNet models do not have any information about the size of the instrument. 
Thus, we use a calibrated camera to capture the final dataset used in this work that sees the instruments from a top-down view with known instrument dimensions. 
We feed this information as metadata to the model to give it more context about the instrument size. 
The equations for the model are given below.
\begin{align}
\mathbf{r} &= f_{\text{res}}^{(R)}(I_R), \mathbf{h} = f_{\text{res}}^{(H)}(I_H), \mathbf{t} = f_{\text{res}}^{(T)}(I_T) \\
\mathbf{m} &= \sigma(W_{m2}\cdot\sigma(W_{m1}\mathbf{x}_m + b_{m1}) + b_{m2}) \\
\mathbf{z} &= [\,\mathbf{r} \,\|\, \mathbf{h} \,\|\, \mathbf{t} \,\|\, \mathbf{m}\,] \\
\hat{\mathbf{y}} &= W_2\cdot\sigma(W_1\mathbf{z} + b_1) + b_2
\end{align}

where $I_R$, $I_H$, and $I_T$ are the cropped image region, head, and tail of the instrument. 
$f_{res}^{(R)}$, $f_{res}^{(H)}$, and $f_{res}^{(T)}$ are the ResNet-18 backbones with the final fully connected (FC) layer removed. $f_{res}^{(H)}$, and $f_{res}^{(T)}$ have shared weights to accommodate flipping and $180 ^\circ$ rotations. 
$\mathbf{x}_m$ is a meta-feature vector of shape $(B,1,2)$ containing the $\log1p$-normalized instrument length and width, where $B$ is the batch size.
$[...\|...]$ denote concatenation. 
$\sigma(.)$ denotes the activation and $\hat{\mathbf{y}}$ denotes the class logits. We used the ReLU activation for all layers and a dropout of 0.2.
This design explicitly enforces part-aware attention and yields robust few-shot discrimination between visually similar instruments where global features are ambiguous.
The Yolo12n was trained for 300 epochs with a batch size of 64 and a starting learning rate of $10^{-2}$ using ultralytics~\cite{yolo12} while the ResNet models were trained for 50 epochs with a batch size of 64 using the Adam optimizer, minimizing the cross entropy loss and a starting learning rate of $10^{-4}$.

\subsubsection{Pose Estimation}
Since the robot will be picking up the surgical instruments and assembling them in the tray, it needs to know the precise location of the instrument on the table, along with its Z rotation ($r_z$). 
To compute this, we use the output of the YOLO detection head to extract the Region of Interest (ROI) for each instrument and feed it into the Segment Anything Model (SAM)~\cite{kirillov2023segment}, which gives us a mask of the instrument as shown in \autoref{fig:pipeline} (segmentation and orientation). 
We then use the contours of the mask and do a Principal Component Analysis~(PCA)~\cite{abdi2010principal} to find the major and minor axes of the instrument. 
The angle between world X and the major axis gives us $r_z$.

\subsection{Manipulation}
\label{sec:manipulation}
\begin{figure}[htb]
    \centering
    \includegraphics[width=\columnwidth]{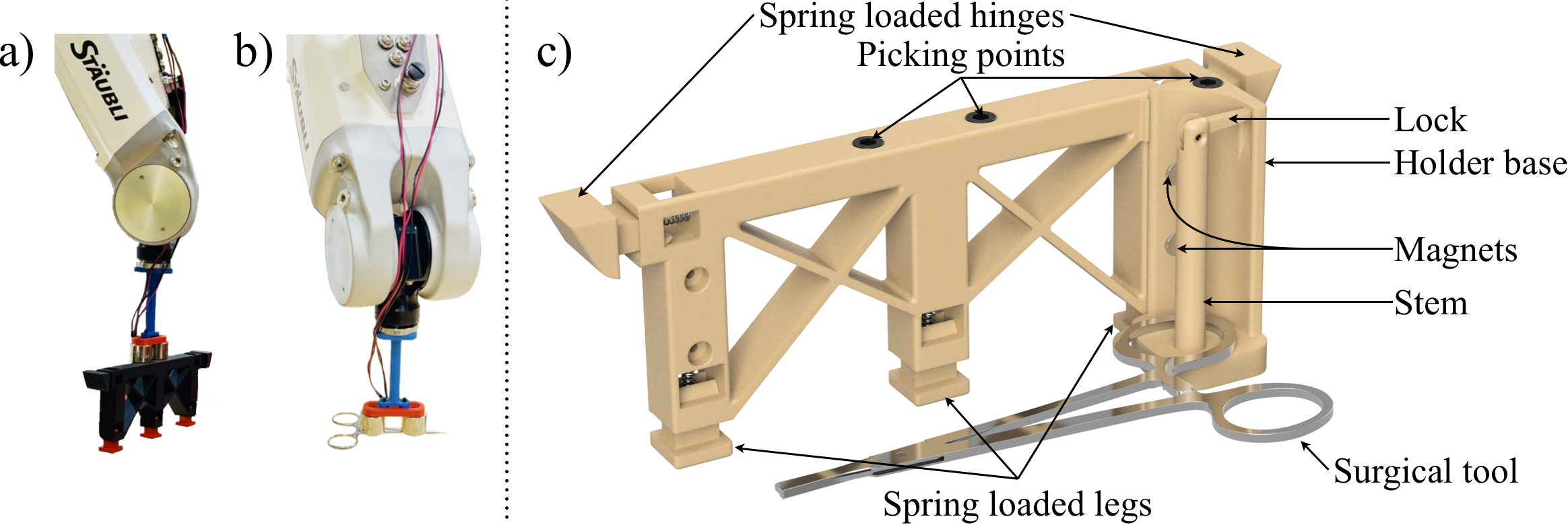}
    \caption{a) CAD render of the divider and holder showing an example instrument; End Effector with the dual electromagnetic picker showing b) divider, c) a straight mosquito forceps}
    \label{fig:picker}
\end{figure}

For picking and placing the instruments, we use the Stäubli TX2-60L 6-DOF robotic arm, which has previously been applied to medical applications and used in hospital environments. 
For the end effector, we custom-designed a dual electromagnetic gripper (shown in \autoref{fig:picker} a and b), controlled by an ESP32 microcontroller. 
The L298-N H-bridge is used to drive the electromagnets. 
We also implemented degaussing in the controller to minimize the effects of the residual magnetic field when the gripper is deactivated. The robot performs the pick and place using a linear point-to-point trajectory.

\subsection{Tray Packing}
\label{sec:packing}
In this section, we discuss the sterile tray packing to reduce collisions and simplify OR preparation.

\subsubsection{Dividers and holders}
\label{sec:divider}
The goal of our system is not only to automate the SPD workflow but also to increase instrument longevity. 
Surgical instruments are often damaged by mechanical impact, friction, or scalding~\cite{tang2024application}. 
To reduce tool-to-tool collision, we designed modular dividers and holders compatible with a wide range of instruments and trays.
\autoref{fig:picker}.c shows the CAD render of the divider with a mounted holder showing a ring-type instrument. 

\begin{figure}[htb]
    \centering
    \includegraphics[width=0.65\columnwidth]{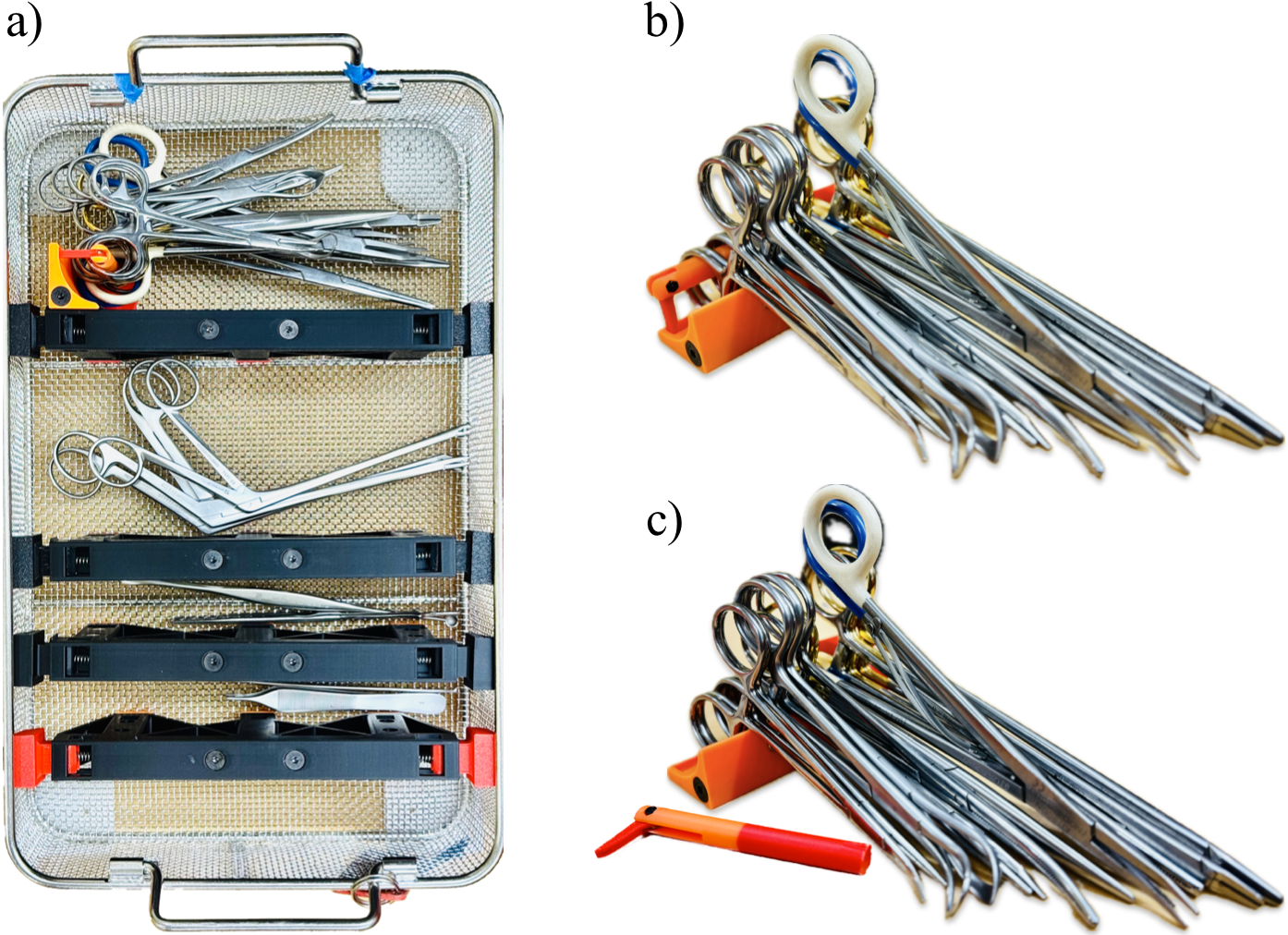}
    \caption{a) Robot assembled tray with 20 different surgical instruments, b) Ring instrument holder, c) and the holder with stem removed}
    \label{fig:tray}
\end{figure}

Each spring-loaded divider locks to the tray brim and uses alignment magnets to position the holder.
The divider is 20 mm thick and has variable height and width. 
The holder consists of a base, a removable stem, and a torsion-spring lock acting as a one-way gate, allowing insertion but preventing instrument fall-out.
In practice, OR nurses would remove the instruments from the sterile tray and place them in order over a folded cloth such that the sides of the instruments are visible at all times. 
The proposed holder can be tilted on the OR table sideways as shown in \autoref{fig:tray} (Right) following OR convention. 
This dual-purpose design simplifies setup and reduces handling time.
All parts are 3D printed with screw-top pick points for the gripper.
The parts are inexpensive to produce and are designed for single-use.

\subsubsection{Algorithm}
\label{sec:algorithm}
To design the packing algorithm efficiently, we interviewed SPD technicians to understand their tray assembly workflow. 
Each technician, for the most part, follows a personal routine of organizing the instruments without standardized guidelines, but most begin by grouping instruments by type. 
Ring-type instruments are held using stringers, while needle and thumb instruments are placed in sterilizer bags. 
Remaining instruments are arranged to reduce movement and collision.
We formalize tray assembly as a ``constrained spatial-allocation" problem and propose a deterministic rule-based packing algorithm that approximates a layout under constraints such as instrument geometry, instrument group, and tray size, which are acquired from the respective datasheets from the manufacturers.
The algorithm~\ref{alg:tool_packing} hierarchically assigns instruments to columns and reduces inter-instrument collision. 
It operates in $O(n)$ time and space.
Dividers separate instrument groups, and the holders discussed in section~\ref{sec:divider} replace stringers.
Instruments are sorted by length within each column, with longer instruments at the bottom.
If dimension constraints are exceeded, the algorithm merges compatible groups in predefined merge sets $M$ - \{Ring, Ring (Thick)\}, \{Ring, Ring (Thick)\} $+$ \{Needle, Thumb\}, \{Ring, Ring (Thick)\} $+$ \{Needle, Thumb, Gun\}.
The robot precomputes instrument, divider, and holder placements before assembly.
The tray, divider, and holders are initially placed at known positions using locking holders on the work desk, making it easier for the technicians to load and unload trays.
The algorithm generalizes across instrument types and tray sizes but currently assumes instrument length is less than tray width and tray aligned orientation, which will be extended in future work.

\begin{algorithm}[H]
\caption{Tray Packing Algorithm}
\label{alg:tool_packing}
\begin{algorithmic}[1]
\Require Instrument quantities $Q$, tray $C$, merge sets $M$, pad $p$
\Ensure Instrument $P$, holder $H$, and divider $D$ placements
\State Initialize $P \gets \emptyset$, $H \gets \emptyset$, $D \gets \emptyset$, $y_{off} \gets 0$
\Statex $C_h, C_w, C_d$ are tray length, width, and depth
\Statex $p_x, p_y, p_z$ are the margins for placement in each axis.
\State Group instruments by merged group $G$ using $M$
\For{each group $g \in G$}
    \State Sort instruments in $g$ by descending length
    \State Compute max instrument length $L_{max}$ and max per layer $N_{max} \gets floor(C_w / (L_{max} + 2p_x))$
    \If{$N_{max} = 0$} \State \textbf{raise} tray width overflow error \EndIf
    \State Initialize $x \gets 0$, $z \gets 0$, staged list $S \gets \emptyset$
    \State Create a queue of all instrument instances from $Q$
    \While{$\vert queue \vert$}
        \State Instruments per layer $n = \min(N_{max}, \vert queue \vert)$
        \If{group is ring type} \State $n \gets 1$ \EndIf
        \State Compute layer height $z_{new}$
        \If{$z_{new} > C_d$}
            \State Commit column $\Rightarrow$ place divider and holders
            \State Reset $S$, $z \gets 0$, continue
        \EndIf
        \For{$i = 1$ to $n$}
            \State Pop instrument $t$, set placement $(x+(i+0.5)(t_l+2p_x), y_{off}, z+t_h)$
            \State Append instance to $S$
        \EndFor
        \State $x \gets 0$, $z \gets z_{new}$
    \EndWhile
    \State Commit remaining column
        \Statex \hspace{1em}Compute column width $w_c$, check height overflow
        \Statex \hspace{1em}Add divider at $(C_w/2, y_{off}+w_c)$ and holder if ring type
        \Statex \hspace{1em}Finalize placements $P$ and update $y_{off} \gets y_{off}+w_c$
\EndFor
\State Remove last divider if beyond tray height
\State \Return $(P, H, D)$
\end{algorithmic}
\end{algorithm}

\section{Evaluation and discussion}
\label{sec:eval}
\subsection{Model performance on classification}
\label{sec:objective}
As discussed in Section~\ref{sec:detection}, initial experiments on a small pilot dataset revealed limitations in YOLO-based classification, motivating the ResNet-based classifier with meta features.
This prior guided the dataset collection in Section~\ref{sec:dataset} and enables the data-scaling analysis presented in this section.

\begin{figure}[htb]
    \centering
    \includegraphics[width=0.95\columnwidth]{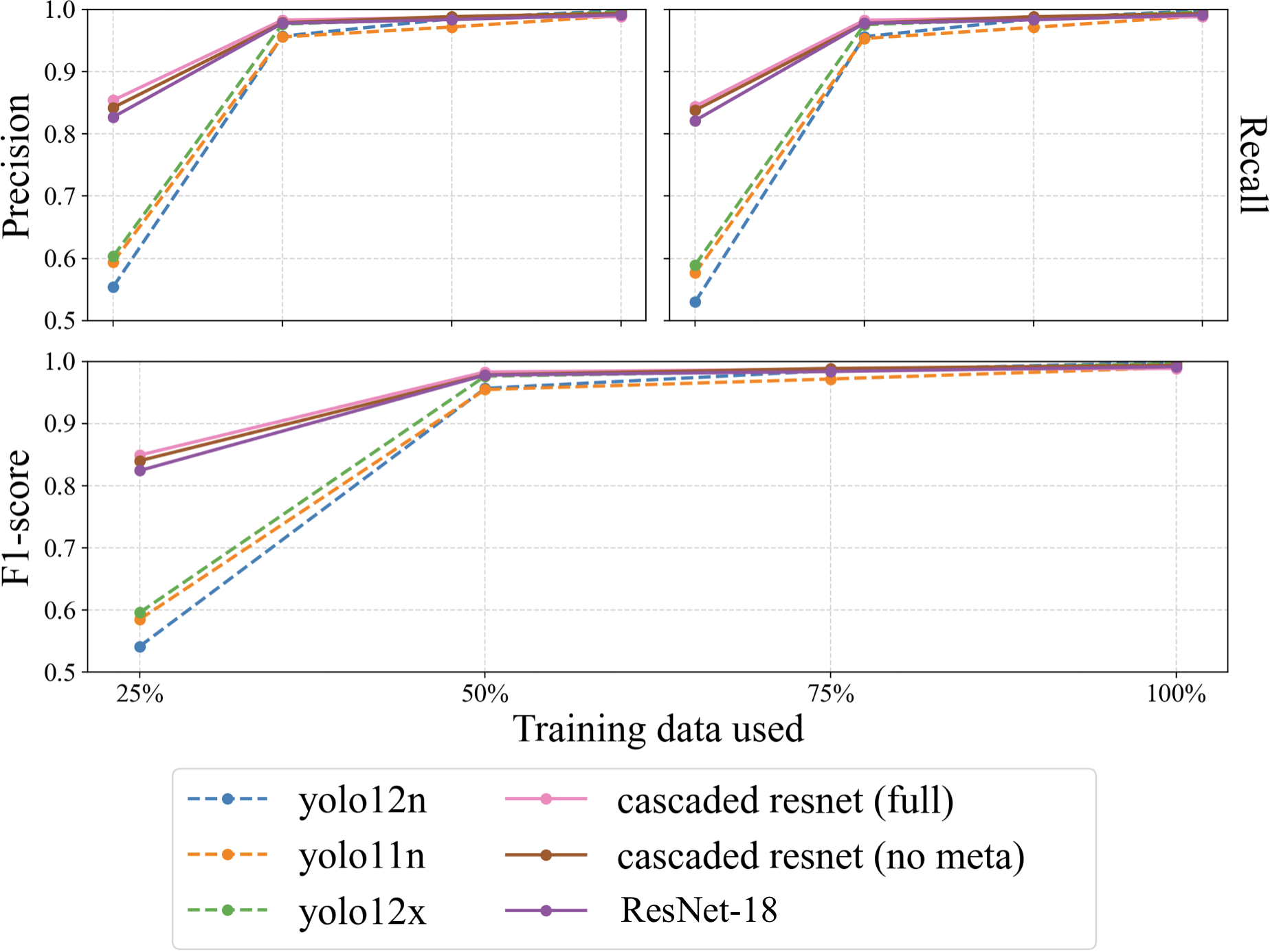}
    \caption{Model performance on the test set vs Training data size}
    \label{fig:data_size_vs_performance}
\end{figure}

We noticed that for this larger dataset, yolo12n performed as good as the ResNet-based model on classification. 
This was an interesting and valuable finding shown in \autoref{fig:data_size_vs_performance} which shows 6 different architectures (3 unified - both detection and classification using the same model and 3 decentralized - detection from yolo12n and classification using the ResNet-based variants) and their classification performance on the test set with varying train data sizes. 
The validation and test sets were kept constant throughout the experiment, while the size of the training data was varied. 
For small data sizes, the ResNet-based decentralized architectures perform significantly better than the unified models. 
As the size of training data increases, unified architectures catch up, and we see a comparable performance on classification.

Another interesting takeaway was that the meta features did not significantly improve the performance, as the F1-score between these models is comparable. 
This could be due to the fact that enough context about the instrument dimensions (in the form of aspect ratio) is already present in the image due to the presence of the black bars (depicted in~\autoref{fig:pipeline} under \textit{region}), as well as due to our cropping the ROI after segmentation, as shown in \autoref{fig:pipeline} - eliminating the need for manually providing the dimensions as meta features.

\subsection{Effects of tray displacement and tilt}
\label{sec:eval:collision}
To empirically assess instrument collision and interference due to the transportation of the sterile tray, a reference (``control") instrument was coated with a fluorescent powder that glows in the presence of ultraviolet (UV) light. 
The control instrument was chosen randomly for each trial. 
After assembling each tray - human technician assembled tray A, tray B manually assembled using dividers and holders but without using the proposed algorithm (i.e., tools within each group are randomly placed) and robot-assembled tray C, we introduced manual vibration for 15 seconds in all 3 axes (X, Y, and Z) with a frequency between $1.5 - 1.8$ Hz measured using high frame rate video recording and a displacement of 100 mm in both directions on each axis, simulating tray transport. 
After vibration, each instrument in the trays was inspected for powder transfer. 
The number of instruments exhibiting transferred powder from the control instrument was recorded as an indicator of collision occurrence. 
The human-assembled tray utilized the stringer to hold the ring group and sterilizer bags, each for thumb and needle groups. 
The robot assembled tray uses the dividers and holders discussed in section~\ref{sec:divider}.

\begin{figure}[htb]
    \centering
    \includegraphics[width=0.8\columnwidth]{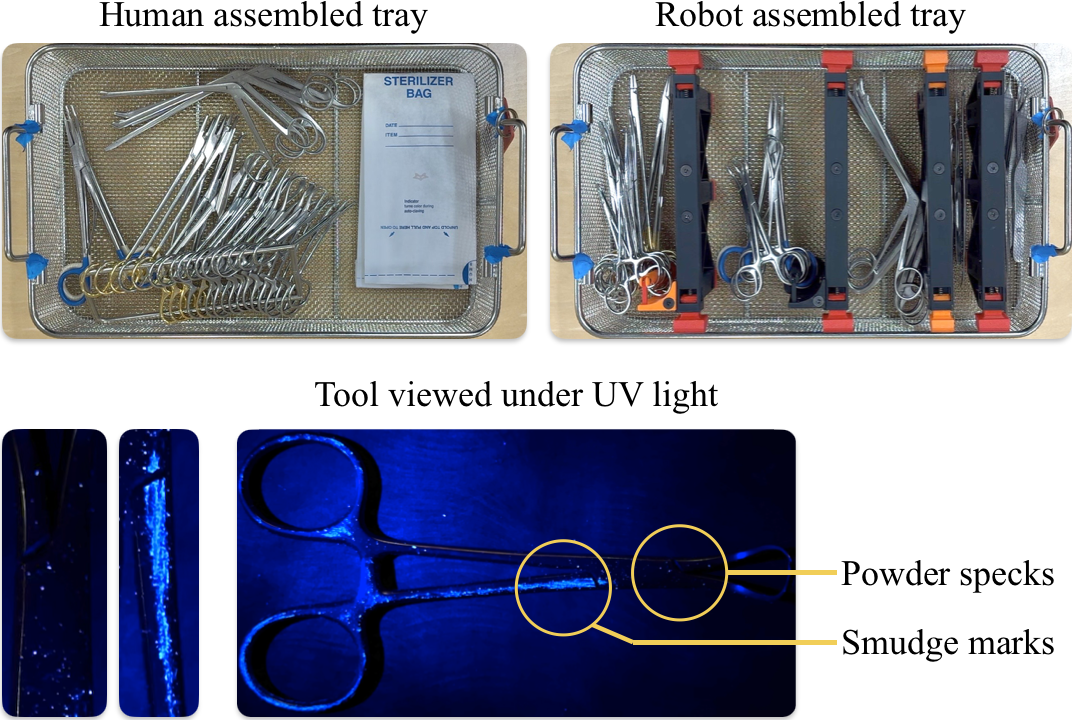}
    \caption{Experimental setup showing human sorted tray and robot sorted tray with an example showing powder specks and smudge marks}
    \label{fig:setup}
\end{figure}

During inspection, any instruments showing powder specks, likely caused by airborne particle deposition, were excluded. 
Such specks can occur due to the spring-back motion of the holder, which can eject fine powder particles into the air during installation. 
Only instruments exhibiting visible smudge marks or continuous powder streaks indicating direct physical contact with the control instrument were considered valid collision events. 
An example of these marks is depicted in \autoref{fig:setup}.

Table~\ref{tab:collision} shows the mean number of instruments that collided in each tray. 
We ran 5 different trials for each tray (A, B, and C), with different control instruments and placements.

To assess instrument collision due to tray tilt, instead of vibrating the tray, we tilted the tray in all 3 axes to about 30 degrees over 3 seconds. 
We ran 5 trials for each tray with different control instruments and placements for this experiment as well.

\begin{table}[htb]
    \centering
    \caption{Instrument collision due to tray displacement and tilt}
    \label{tab:collision}
    \begin{tabular}{|c|c|c|}
        \hline
        & \multicolumn{2}{c|}{\textbf{mean collision (std.)} \textit{- lower is better}} \\ \hline
        \textbf{Tray} & \textbf{Displacement} & \textbf{Tilt} \\ \hline
        baseline (A) & 11.4 (5.004) & 7.8 (3.763) \\
        No Algorithm (B) & 5.0 (2.0) & 3.6 (1.02) \\
        proposed (C) & \textbf{4.0 (1.096)} & \textbf{1.8 (0.4)} \\ \hline
    \end{tabular}
\end{table}

Compared to the baseline, the proposed method significantly reduced collisions under both displacement (Cohen's~$d = 2.04$) and tilt ($d = 2.24$), corresponding to very large effect sizes. 
While statistically significant ($p < 0.05$), the limited number of trials motivates further validation in a large-scale and real SPD setting.  
Due to the inclusion of holders and dividers, instrument groups remained physically isolated within their respective columns, and no inter-column contact was observed under either displacement or tilt conditions. 
Consequently, in most trials, the control instrument exhibited collisions only with the immediate neighboring instrument located above or below it within the column.
Furthermore, the hierarchical stacking strategy contributed to enhanced mechanical stability, thereby reducing instrument movement and collision frequency. 
This effect was validated through (B), in which instruments were placed randomly within each column, demonstrating a higher rate of collision compared to the hierarchical arrangement.

These experiments provide the first evidence that structured robotic packing can mitigate collision-induced wear, an important contributor to instrument degradation. 
However, it does not assess human factors such as perceived ease of use or the cognitive load. 
We plan to conduct a user study with OR nurses and SPD technicians in the future to quantify perceived ease, instrument retrieval time, and workload.

\section{Conclusion and Future works}
\label{sec:future_work}
In this work, we demonstrated a clinically relevant robotic system for automatic sorted assembly of pre-op surgical instruments in sterile trays. 
The system integrates computer vision, robotics, and customized parts to automate one of the most time-consuming and error-prone stages of sterile processing. 
Through a custom dataset and an end-to-end perception-manipulation pipeline, the robot achieves consistent localization, precise placement, and structured assembly of instruments within the tray while interacting with and aiding human technicians in the workflow. 
Our results demonstrate a substantial reduction in instrument collisions and displacement compared to manually assembled trays, validating the system’s robustness and potential for real-world deployment in SPD units.

In the future, we aim to conduct a user study with OR nurses and SPD technicians to evaluate the perceived usability, efficiency, and cognitive load. 
We also look forward to expanding the dataset with more instrument classes and larger sample sizes to improve generalization across different hospital environments. 
We plan to extend the packing algorithm to multi-orientation and multi-layer layouts using cost-based planning.
We aim to improve the picker design to handle non-magnetic instruments (such as titanium or plastics) and also to be able to manipulate instrument closing and opening, which are part of the standard SPD workflow.

\section*{Acknowledgment}
We sincerely thank Rawan Ahmed and Harshith Jella for their support in collecting and annotating the dataset.
\bibliographystyle{IEEEtran}
\bibliography{ref}
\end{document}